\documentclass[journal]{IEEEtranTIE}
\usepackage{graphicx}
\usepackage{cite}
\usepackage{picinpar}
\usepackage{amsmath}
\usepackage{url}
\usepackage{flushend}
\usepackage[utf8]{inputenc}
\usepackage{colortbl}
\usepackage{soul}
\usepackage{multirow}
\usepackage{pifont}
\usepackage{color}
\usepackage{alltt}
\usepackage{enumerate}
\usepackage{siunitx}
\usepackage{breakurl}
\usepackage{epstopdf}
\usepackage{pbox}

\usepackage{tabu}                     % 表格插入
\usepackage{multirow}                 % 一般用以设计表格，将所行合并
\usepackage{multicol}                 % 合并多列
\usepackage{multirow}                % 合并多行
\usepackage{float}                    % 图片浮动
\usepackage{makecell}                 % 三线表-竖线
\usepackage{booktabs}                 % 三线表-短细横线

\usepackage{graphicx}
\usepackage{cite}
\usepackage{picinpar}
\usepackage{amsmath}
\usepackage{url}
\usepackage{flushend}
\usepackage{colortbl}
\usepackage{soul}
\usepackage{multirow}
\usepackage{pifont}
\usepackage{color}
\usepackage{alltt}
\usepackage{enumerate}
\usepackage{siunitx}
\usepackage{breakurl}
\usepackage{epstopdf}
\usepackage{pbox}
\usepackage[vlined,ruled]{algorithm2e}
\usepackage{caption}
\usepackage{subcaption}
\usepackage{booktabs} % 用于绘制三线表格
\usepackage{multirow}  % 合并多行
\usepackage{xcolor}
\usepackage{threeparttable}
\usepackage{graphics}
\usepackage{amssymb}
\usepackage{booktabs}
\usepackage{graphicx}
\usepackage{cite}
\usepackage{picinpar}
\usepackage{amsmath}
\usepackage{url}
\usepackage{flushend}
\usepackage{colortbl}
\usepackage{soul}
\usepackage{multirow}
\usepackage{pifont}
\usepackage{color}
\usepackage{alltt}
\usepackage{enumerate}
\usepackage{siunitx}
\usepackage{breakurl}
\usepackage{epstopdf}
\usepackage{pbox}
\usepackage[vlined,ruled]{algorithm2e}
\usepackage{caption}
\usepackage{subcaption}
\usepackage{booktabs} % 用于绘制三线表格
\usepackage{multirow}  % 合并多行
\usepackage{xcolor}
\usepackage{threeparttable}
\usepackage{graphics}
\usepackage{amssymb}
\usepackage{booktabs}
\usepackage[hidelinks,colorlinks,bookmarksopen,bookmarksnumbered,citecolor=green, linkcolor=red, urlcolor=green]{hyperref}

\begin{document}
\title{	DOA: A Degeneracy Optimization Agent with Adaptive Pose Compensation Capability based on Deep Reinforcement Learning}

\author
{
	\vskip 1em
	
	Yanbin Li, Canran Xiao, Hongyang He, Shenghai Yuan, Zong Ke, Jiajie Yu, Zixiong Qin, 
    Zhiguo Zhang, \\ Wenzheng Chi* and Wei Zhang* 

	\thanks
{
	
    This project was supported by National Natural Science Foundation of China grant \#62273246 awarded to Wenzheng Chi and by the Fund of State KeyLaboratory of IPOC (BUPT) under Grant IPOC2025ZT07.

Yanbin Li, Wei Zhang, Zixiong Qin and Zhiguo Zhang are with School of Electronics Engineering, Beijing University of Posts and Telecommunications, Beijing, 100876, China (yanbinli@bupt.edu.cn; weizhang13@bupt.edu.cn; zxqin@bupt.edu.cn; zhangzhiguo@bupt.edu.cn;).

Canran Xiao  is with Xiangjiang Laboratory, Changsha, 410205, China (xiaocanran@csu.edu.cn). 
Hongyang He is with Signal and Information Processing (SIP) Lab, Department of Computer Science, University of Warwick, Coventry, CV4 7AL, UK (Hongyang.He@warwick.ac.uk).
Shenghai Yuan is with the School of Electrical and Electronic Engineering, Nanyang Technological University, 50 Nanyang Avenue, Singapore 639798 (shyuan@ntu.edu.sg).
Zong Ke is with the Department of Statistics and Probability, National University of Singapore, 21 lower kent ridge road, Singapore 119077 (a0129009@u.nus.edu).

Wenzheng Chi and Jiajie Yu are with Robotics and Microsystems Center, School of Mechanical and Electric Engineering, Soochow University, Suzhou, 215021, Jiangsu, China (wzchi@suda.edu.cn; jjyu986@stu.suda.edu.cn).

    $^{*}$ corresponding author
}

}

\maketitle

\begin{abstract}

Particle filter-based 2D-SLAM is widely used in indoor localization tasks due to its efficiency. However, indoor environments such as long straight corridors can cause severe degeneracy problems in SLAM. In this paper, we use Proximal Policy Optimization (PPO) \cite{schulman2017proximalpolicyoptimizationalgorithms} to train an adaptive degeneracy optimization agent (DOA) to address degeneracy problem. 
We propose a systematic methodology to address three critical challenges in traditional supervised learning frameworks: (1) data acquisition bottlenecks in degenerate dataset, (2) inherent quality deterioration of training samples, and (3) ambiguity in annotation protocol design.
We design a specialized reward function to guide the agent in developing perception capabilities for degenerate environments. Using the output degeneracy factor as a reference weight, the agent can dynamically adjust the contribution of different sensors to pose optimization. Specifically, the observation distribution is shifted towards the motion model distribution, with the step size determined by a linear interpolation formula related to the degeneracy factor. In addition, we employ a transfer learning module to endow the agent with generalization capabilities across different environments and address the inefficiency of training in degenerate environments.
Finally, we conduct ablation studies to demonstrate the rationality of our model design and the role of transfer learning. We also compare the proposed DOA with SOTA methods to prove its superior degeneracy detection and optimization capabilities across various environments. 
The source code and supplementary materials will be available at: https://github.com/littleBurgerrr/DOA\_slam.git.

\end{abstract}


% \markboth{IEEE TRANSACTIONS ON INDUSTRIAL ELECTRONICS}%
% {}

\definecolor{limegreen}{rgb}{0.2, 0.8, 0.2}
\definecolor{forestgreen}{rgb}{0.13, 0.55, 0.13}
\definecolor{greenhtml}{rgb}{0.0, 0.5, 0.0}

\begin{figure}[t]
\centerline{\includegraphics[width=0.5\textwidth]{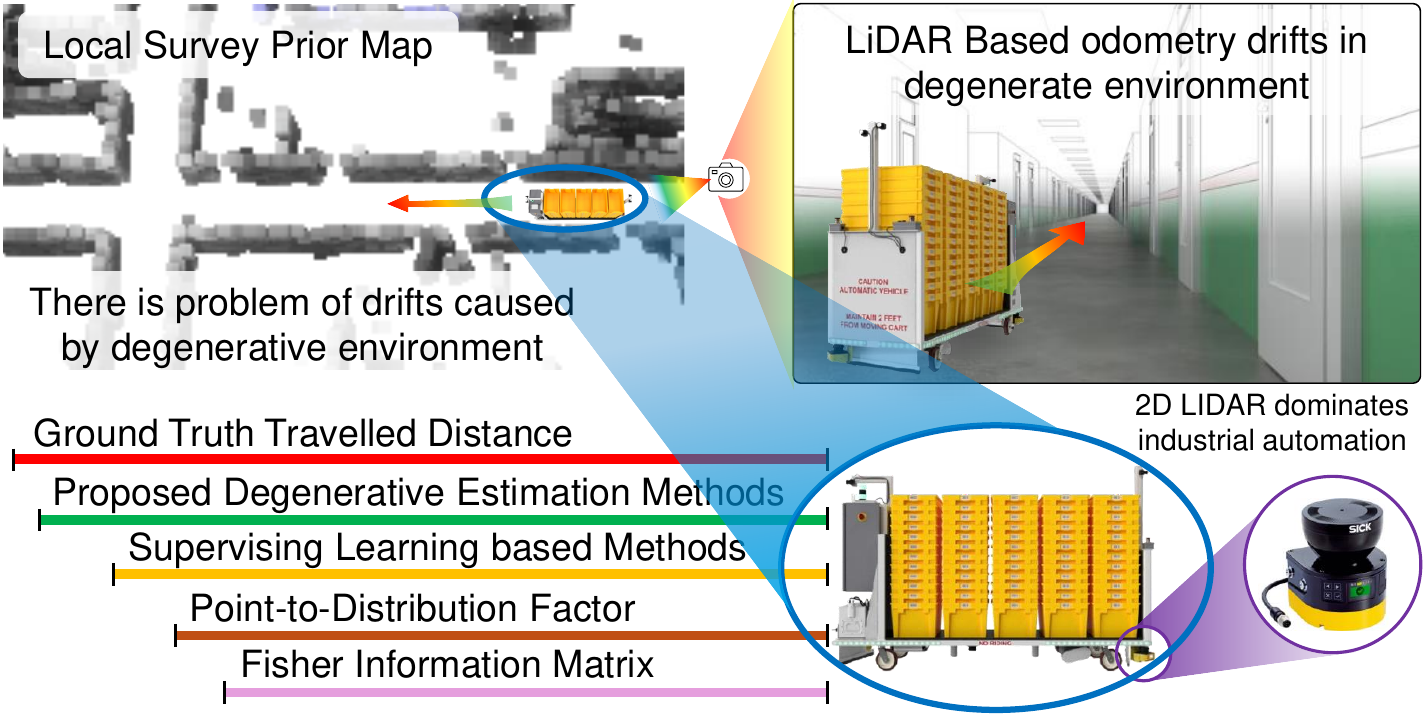}}

% \caption{The degeneracy problem of Lidar-SLAM occurs in the long straight corridor. Degeneracy detection accuracy directly affects localization accuracy. The error between our method and the ground truth is the smallest among all the methods.}
\caption{The degeneracy problem in Lidar-SLAM is particularly pronounced in long straight corridors, where the lack of geometric features can lead to estimation errors. The accuracy of degeneracy detection directly impacts localization precision. Among all the methods compared, our method exhibits the smallest error relative to the ground truth (GT).}

\label{DOA_overview}
\end{figure}

\section{INTRODUCTION}

\IEEEPARstart{S}{imultaneous} localization and mapping (SLAM) aims to address the problem of localization and mapping for robots, which is a prerequisite for autonomous navigation in unknown environments. 2D-SLAM has been widely used in indoor localization for robots due to low cost and efficiency \cite{grisetti2007improved, 7487258, kohlbrecher2011flexible}. GMapping \cite{grisetti2007improved}, based on particle filter, works well in indoor scenes with high real-time requirements. It is widely used in robotic vacuum cleaners and industrial robots, because its two-dimensional maps are suitable for floor-cleaning and industrial transportation tasks.

However, indoor tasks often encounter degenerate environments (geometry feature-less environments like long straight corridor) that causes localization drift. In degenerate environments, camera-based SLAM lacks sufficient visual features \cite{campos2021orb, 10740921, 10305271, 10874215}. These environments often contain large homogeneous regions with no distinct textures or structures, making it difficult to extract features to track the position.
Similarly, lidar data lacks distinct structural features \cite{zhang2014loam, shan2018lego}, resulting in uniform depth information and insufficient constraints for pose estimation.
In degenerate environments, lidar and cameras often fail, while odometry can still provide short-term initial pose estimates despite its cumulative error. However, tightly-coupled multi-sensor fusion methods with fixed rules, cannot dynamically adjust sensor contributions to pose estimation and fail to achieve degeneracy optimization \cite{10757429, 10404014, 10776572, 10631284, lee2024switch}.

The key challenge of degeneracy optimization (eliminate degeneracy problem caused by the lack of constraints in the optimization of SLAM) is accurately detecting degeneracy and applying dynamic compensation based on the degree of degeneracy using constraints from different sensors. As shown in Fig. \ref{DOA_overview}, traditional methods based on fixed rules rely too much on preset thresholds and parameter tuning \cite{10816047,zhou2020lidar}, supervised learning methods depend heavily on dataset quality and face issues with ambiguous annotation rules \cite{nubert2022learning,10994466}, these limitations reduce detection accuracy and prevent timely optimization, leading to localization drift.

To address the degeneracy problem in particle filter-based 2D-SLAM, we train an agent using PPO. Agent learns optimal policy through a specialized reward and interaction with environment, overcoming the dependence on the quality of dateset annotations. The agent is capable of analyzing particle distributions to detect degeneracy and dynamically adjust the contribution weights of different sensors to pose optimization.
Specifically, we fine-tune the failed observation distribution towards the motion model distribution by shifting it along the line connecting their centroids. The shift magnitude is determined by a linear interpolation based on the degeneracy factor output by agent. Finally, we select the optimal pose by comparing the likelihood scores of the updated and original distributions.
Additionally, we utilize transfer learning to preserve the learned capabilities of agent for general features within the lower-level network and extend to other environments. This approach enhances the generalization ability and training performance of agent. To our knowledge, this is the first work using PPO to achieve dynamic sensor fusion to improve particle filter-based SLAM performance in degenerate environments. The primary contributions are:

\begin{enumerate}[1)]
	\item A PPO-based online simulation training framework equips the agent with dynamic perception ability for degenerate environments, enabling adaptive sensor fusion for optimization. It addresses challenges in traditional supervised learning, such as difficulties in collecting high-quality datasets and ambiguous labeling rules;
    
	\item A transfer learning module which leverages the learned capabilities for general features to enhance the decision-making ability of agent across diverse environments and improve training efficiency in degenerate scenes;

    \item Feasibility and performance of system are verified by mathematical proof, ablations and comparative experiments.
	
\end{enumerate}

\section{RELATED WORK}

\subsection{Degeneracy Detection}

Degeneracy detection is usually the pre-process of degeneracy optimization.
Chen $\textit{et al.}$ \cite{10816047} judged the degeneracy degree by observing the distribution probability changes of voxelized point clouds. However, this relies on predefined thresholds and assumes a Gaussian distribution of point clouds.
Zhou $\textit{et al.}$ \cite{zhou2020lidar} detected degeneracy by analyzing the eigenvalues of Fisher information matrix from lidar and ultra wide band (UWB), but threshold selection is still subjective.
Lee $\textit{et al.}$ \cite{lee2024switch} analyzed the eigenvalues of Hessian matrix for degeneracy detection and defined a non-heuristic threshold using chi-square test. However, the detection accuracy depends on the confidence level of the chi-square test and requires parameter tuning in different environments.
There are some works that use networks to capture feature changes during data interactions, eliminating the dependency on thresholds in traditional methods.
Nubert $\textit{et al.}$ \cite{nubert2022learning} trained a network on simulated data to predict the localizability of lidar data. 
Li $\textit{et al.}$ \cite{10994466} trained a model to perceive changes in the distribution features of particle swarms to achieve degeneracy detection. However, these learning-based methods heavily rely on the quality of dataset. The ambiguity of manual labeling rules makes it difficult for traditional supervised learning to accurately capture subtle signs of degeneracy.
This paper addresses these issues by dynamically adjusting the behavior of agent through dynamic interaction and adaptive reward mechanism of deep reinforcement learning (DRL).

\subsection{Degeneracy Optimization \& DRL-based Methods}

Tuna $\textit{et al.}$ \cite{tuna2023x} proposed a lidar registration method, introducing additional constraints for iterative closest point (ICP) based on the results of localizability detection. 
Chen $\textit{et al.}$ \cite{10816047} weighted point clouds using the degeneracy factor from P2d to optimize pose estimation. However, these rule-based methods generalize poorly across environments.
Torroba $\textit{et al.}$ \cite{torroba2020pointnetkl} used a model to predict the covariance matrix of point cloud registration as measurement weights for degeneracy optimization.
Landry $\textit{et al.}$ \cite{landry2019cello} uses data-driven method to predict ICP covariance for pose estimation. However, these methods relies heavily on the accuracy of prediction.

Regarding the application of DRL, Botteghi $\textit{et al.}$ \cite{botteghi2020reinforcement} proposed a reinforcement learning (RL) approach to enhance SLAM by learning to build maps, introducing a reward mechanism to optimize map-building process.
Messikommer $\textit{et al.}$ \cite{messikommer2024reinforcement} proposed a RL approach to enhance visual odometry by predicting camera motion, improving accuracy in challenging environments.
Inspired by the use of DRL to improve SLAM modules, in this paper, we use PPO to guide agent behavior, where the results of degeneracy optimization serve as feedback, improving robustness in various environments.

We will demonstrate the feasibility of our method in the next section and provide mathematical model and formula proof.

\begin{figure*}[ht]
    \centerline{\includegraphics[width=0.99\textwidth]{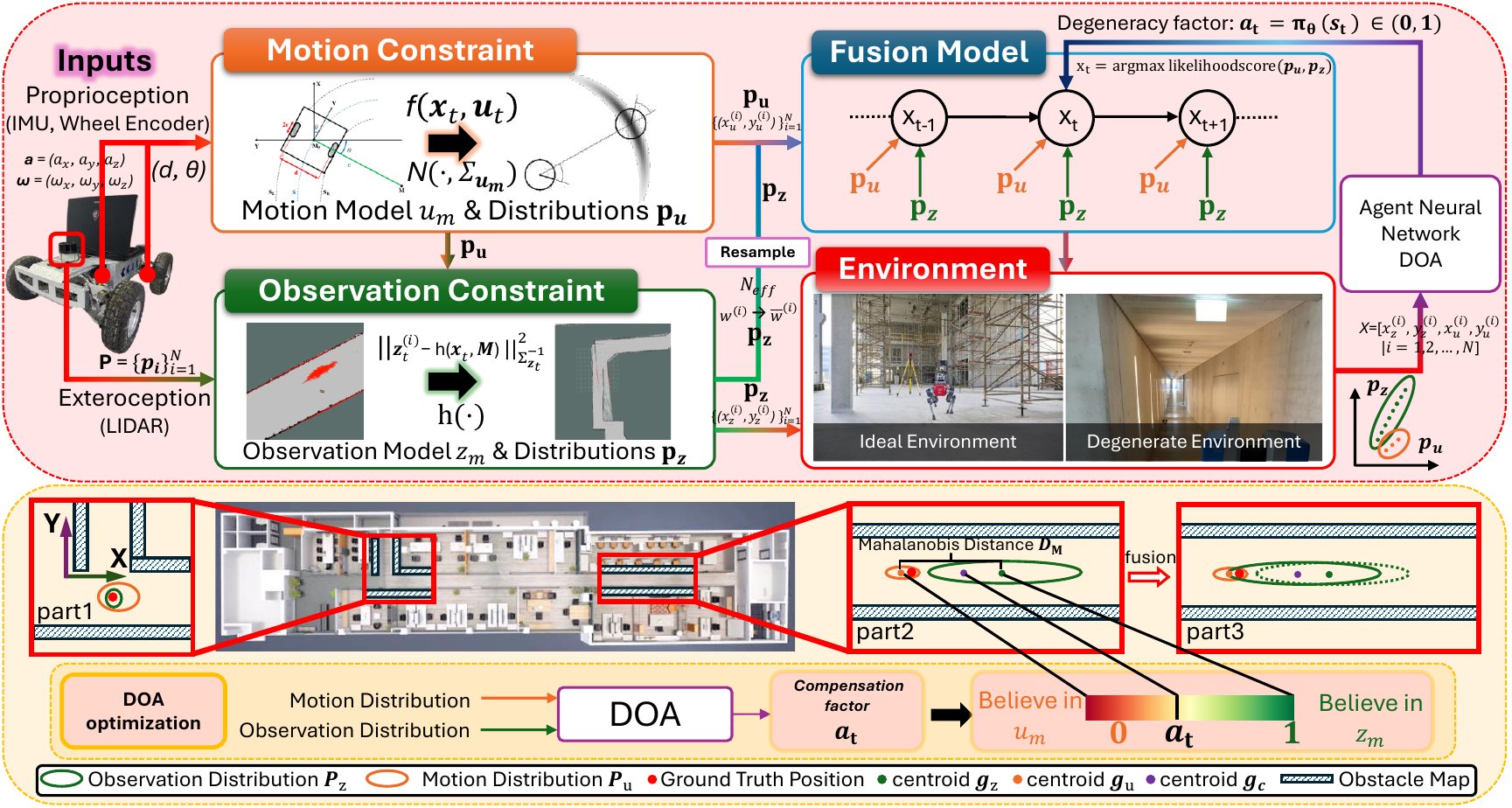}}
  
    \caption{The upper part is overall process. Lack of observation constraint makes red particle point set diverge, reducing localization accuracy and leading to the failure of map loop closure. Our method compensates $\textbf{p}_z$ and $\textbf{p}_u$ using the factor output by DOA through fusion model. The lower part is detailed demonstration of fusion model with DOA. Part1, part2 and part3 respectively demonstrate the states of different distributions in normal environment, degenerate environment and after DOA optimization.}
  
    \label{pipeline}
\end{figure*}

\section{PRELIMINARY}

Particle filter-based SLAM estimates the state of robot through a set of particles, each particle carries coordinates $(x,y) \in \mathbb{R}^{2}$ and orientation angle $\theta$. The joint posterior probability problem of SLAM can be decomposed into two parts: localization and mapping.
\begin{equation}
\label{divide2}
p(\textbf{x}_{t},\textbf{M}\vert \textbf{z}_{t},\textbf{u}_{t})=p(\textbf{M}\vert \textbf{x}_{t},\textbf{z}_{1:t})\cdot p(\textbf{x}_{t}\vert \textbf{z}_{t},\textbf{u}_{t})    
\end{equation}
where $\textbf{x}_{t}\in \mathbb{R}^{3}$ is optimal pose of robot (the one with the highest scan matching score among particle swarm poses) at time $t$, $\textbf{z}_{t}\in \mathbb{R}^{n}$ and $\textbf{u}_{t}\in \mathbb{R}^{3}$ are lidar observation data and the input of motion model $u_m$ respectively, $\textbf{M}$ is global map.

Localization accuracy impacts map quality by propagating pose errors to map features through observation model $z_m$ (mathematical model for optimizing pose based on lidar data), inflating covariance matrix $\Sigma_{\textbf{M}_{k}}\in \mathbb{R}^{2\times2}$ and increasing its uncertainty, so we mainly focus on localization accuracy. Formulas for $z_m$ and $\Sigma_{\textbf{M}_{k}}$ are in (\ref{observation}) and (\ref{cov_map}).
\begin{equation}
\label{observation}
z_{m} = h(\textbf{x}_{t} , \textbf{M}_{k}) + \varepsilon _ { t } \sim N ( 0 , \Sigma _ { \textbf{z} _t} )
\end{equation}
where $h(\cdot)$ is observation function, $\textbf{M}_{k}$ is the $k_{th}$ feature point in global map $\textbf{M}$, $\varepsilon_t$ is Gaussian noise and $\Sigma_{\textbf{z}_t}$ is covariance matrix of observation data $\textbf{z}_t$.
\begin{equation}
\label{cov_map}
\Sigma_{ \textbf{M}_{k}} = J_{h}^{-1}( \Sigma_{z}+J_{\textbf{x}_t} \Sigma_{\textbf{x}_t} J_{\textbf{x}_t}^{T}) (J_{h}^{-1})^{T}
\end{equation}
where $\Sigma_{\textbf{x}_t}$ is covariance matrix of $\textbf{x}_t$, $J_{h}$ and $J_{\textbf{x}_t}$ are Jacobian matrices of $h(\cdot)$ and $\textbf{x}_t$ respectively.

The pose optimization of particle filter-based SLAM is divided into two parts. 
First, motion model $u_m$ shown in (\ref{motion_model}) initially predicts pose based on odometry information to obtain motion distribution $\textbf{p}_u$ (the orange point set composed of particle swarm pose in Fig. \ref{pipeline}). 
Subsequently, observation model $z_m$ use scan matching in (\ref{ob_model}) to further optimize pose to obtain observation distribution $\textbf{p}_z$ (the green point set composed of particle swarm pose in Fig. \ref{pipeline}). 
As shown in (\ref{scan_match}), scan matching optimizes pose by aligning lidar scan with map to correct pose $(x_z,y_z)\in \mathbb{R}^{2}$, and then calculate likelihood score $s^{(i)}\in \mathbb{R}$ for each particle.
\begin{equation}
\label{motion_model}
p(\textbf{x}_{t}\vert \textbf{x}_{t-1},\textbf{u}_{t})=N(f(\textbf{x}_{t-1},\textbf{u}_{t}),\Sigma_{u_m})
\end{equation}
% where $\textbf{u}_t=[\Delta x, \Delta y, \Delta \theta ]^T$ is odometry increment,
where $f(\cdot)$ is nonlinear motion equation, and $\Sigma_{u_m}$ is covariance matrix of odometry noise.
\begin{equation}
\label{ob_model}
p(\textbf{z}_{t}\vert \textbf{x}_{t},\textbf{M}) \propto exp(-\frac{1}{2}\sum_{i=1}^{n}\vert \vert \textbf{z}_{t}^{(i)}-h(\textbf{x}_{t},\textbf{M})\vert \vert _{\sum_{\textbf{z}_t}^{-1}}^{2})
\end{equation}
where $\textbf{z}_{t}^{(i)}$ is the $i_{th}$ measurement in laser beam of total $n$, and $\Sigma_{\textbf{z}_t}$ is covariance matrix of observation noise.
\begin{equation}
\label{scan_match}
(x_z,y_z) = \arg \min_{\textbf{x}_{t}}\sum_{i=1}^{n}\vert \vert \textbf{z}_{t}^{(i)}-h(\textbf{x}_{t},\textbf{M})\vert \vert _{\sum_{\textbf{z}_t}^{-1}}^2
\end{equation}

In normal environment of Fig. \ref{pipeline} (part1), robot obtains accurate observation constraints, making $\textbf{p}_z$ relatively concentrated and reliable (more concentrated distribution $\rightarrow$ smaller sampling area $\rightarrow$ better pose accuracy). The intersection area between $\textbf{p}_u$ and $\textbf{p}_z$ represents the optimal sampling region for pose estimation.

In degenerate environment in Fig. \ref{pipeline} (part2), robot moves along x-axis (the coordinate system has been given in Fig. \ref{pipeline}), the lack of constraints causes the component of Jacobian matrix $H_{\textbf{z}_t}=\left[\frac{\partial h}{\partial x} , \frac{\partial h}{\partial y},\frac{\partial h}{\partial \theta}\right]$ to approach zero in degenerate direction. 
As shown in (\ref{eq2}), the information matrix $I_{z_m}=J_h^T\Sigma_{\textbf{z}_t}^{-1}J_h$ of $z_m$ becomes rank-deficient in this direction. Consequently, the total information matrix $I(\textbf{x}_{t})=I_{z_m} + I_{u_m}$ in (\ref{eq3}) has insufficient constraints in the x-direction. 
\begin{equation}
\label{eq2}
I_{z_m} \approx \begin{bmatrix}
\lambda_{x} & 0 & 0 \\
0 & \lambda_{y} & 0 \\
0 & 0 & \lambda_{\theta}
\end{bmatrix}, \quad \lambda_{x} \rightarrow 0
\end{equation}
\begin{equation}
\label{eq3}
I(\textbf{x}_{t}) \approx \begin{bmatrix}
 (I_{u_m})_x& 0 & 0 \\
0 & \lambda_{y}+  (I_{u_m})_y  & 0 \\
0 & 0 & \lambda_{\theta}+  (I_{u_m})_\theta
\end{bmatrix}
\end{equation}
where $I_{u_m}$ is information matrix of $u_m$, $(I_{u_m})_x$, $(I_{u_m})_y$ and $(I_{u_m})_\theta$ are three diagonal components of $I_{u_m}$ respectively.
Then it reflect covariance matrix $\Sigma_{\textbf{x}_{t}}$, the diagonal elements $(\sigma_{x}^{2},\sigma_{y}^{2},\sigma_{\theta}^{2})$ of $\Sigma_{\textbf{x}_{t}}$ respectively control the accuracy of pose components $(x,y,\theta)$. 
Due to the lack of constraints along x-axis and limited constraints along y-axis caused by long-term motion in a single direction, both $\sigma_{x}^{2}$ and $\sigma_{y}^{2}$ tend to increase significantly, with $\sigma_{x}^{2}$ diverging more severely in (\ref{eq4}).
\begin{equation}
\label{eq4}
\sigma_{x}^{2}=I^{-1}(\textbf{x}_t)\approx\frac{1}{(I_{u_m})_{x}} \longrightarrow \infty
\end{equation}

In this case, the optimization algorithm cannot effectively update the pose in x-direction, leading to the Hessian matrix $H=\frac{\partial^{2}E}{\partial x^{2}}$ of objective function shown in (\ref{eq5}) being singular in x-direction, where $h$ is second derivative.
\begin{equation}
\label{eq5}
H \approx \begin{bmatrix}
0 & 0 & 0 \\
0 & h_{yy} & h_{y\theta} \\
0 & h_{\theta y} & h_{\theta\theta}
\end{bmatrix}
\end{equation}
At the same time, scan matching fail and lead to low likelihood scores $s^{(i)}$. This manifests as $\textbf{p}_z$ diverge along x-axis, stretching particle distribution. As shown in Fig. \ref{pipeline} (part2), the divergence of $\textbf{p}_z$ on x-axis is particularly obvious and accompanied by a drift phenomenon (while $\textbf{p}_u$ is relatively more reliable, GT pose is within its sampling area). And Mahalanobis distance $D_M \in \mathbb{R}$ between $\textbf{p}_u$ (with centroid $\textbf{g}_u = (x_u, y_u)$ = mean value of coordinates) and $\textbf{p}_z$ (with centroid $\textbf{g}_z = (x_z, y_z)$) is positively correlated with the drift distance of $\textbf{p}_z$ and the degree of degeneracy:
% Additionally, it means particles unaffected by degeneracy should exhibit clustering.
\begin{equation}
D_{M}=\sqrt{({x}_{u}-{x}_{z})^{T}\Sigma_{\textbf{x}_{t}}^{-1}({x}_{u}-{x}_{z})}
\end{equation}
At this point, the optimal pose lies along the line connecting $\textbf{p}_u$ and $\textbf{p}_z$. Since $\textbf{p}_u$ is more reliable than $\textbf{p}_z$ in degenerate environments, we fuse $\textbf{p}_u$ and $\textbf{p}_z$ to integrate multi-sensor information, thereby obtaining the fused distribution $\textbf{p}_c$ (with centroid $\textbf{g}_c = (x_c, y_c)$) by optimizing $\textbf{p}_z$. We balance contributions of $\textbf{p}_u$ and $\textbf{p}_z$ with weight $a \in \mathbb{R}$ in fusion formula:
\begin{equation}
\begin{split}
&\Sigma_{c}^{-1} = (1-a) \Sigma_{z}^{-1}+a\Sigma_{u}^{-1}, \\
&\textbf{g}_{c}=\Sigma_{c}\left((1-a) \Sigma_{z}^{-1}\textbf{g}_{z}+a\Sigma_{u}^{-1}\textbf{g}_{u}\right)
\end{split}
\end{equation}
where $\Sigma_{u}^{-1}$, $\Sigma_{z}^{-1}$ and $\Sigma_{c}^{-1}$ are covariance matrices of $\textbf{p}_u$, $\textbf{p}_z$ and $\textbf{p}_c$ respectively.
In degenerate environment, the information matrix $\Sigma_z^{-1}$ of $\textbf{p}_z$ approaches zero in degenerate directions. Consequently, $\Sigma_c^{-1}$ and $\textbf{g}_c$ degenerate into:
\begin{equation}
\begin{split}
&\Sigma_{c}^{-1}\approx a\Sigma_{u}^{-1}, \\
&\textbf{g}_{c}\approx \Sigma_{c}(a\Sigma_{u}^{-1}\textbf{g}_{u})=\textbf{g}_{u}
\end{split}
\end{equation}
This indicates that fusion result is dominated by $\textbf{p}_u$, and optimal fused mean $\textbf{g}_c$ can be obtained by solving the minimization problem in (\ref{optimization}):
\begin{equation}
\begin{split}
% \textbf{g}_c =(1-a) \cdot \textbf{g}_z + a \cdot \textbf{g}_u 
&\textbf{g}_c = \arg\min_{g} \left( (1 - a)\|\textbf{g} - \textbf{g}_z\|^2 + a\|\textbf{g} - \textbf{g}_u\|^2 \right),\\
&solution:\textbf{g}_c =(1-a) \cdot \textbf{g}_z + a \cdot \textbf{g}_u 
\end{split}
\label{optimization}
\end{equation}

Thus, we have demonstrated the feasibility of the method. In degenerate optimization phase in Fig. \ref{pipeline} (part3), we globally fine-tune $\textbf{p}_z$ toward $\textbf{p}_u$, translating $\textbf{g}_z$ along the line between $\textbf{g}_z$ and $\textbf{g}_u$ until it coincides with $\textbf{g}_c$. We will describe the agent modeling in the next section.

\begin{figure*}[t]
\centerline{\includegraphics[width=0.95\textwidth]{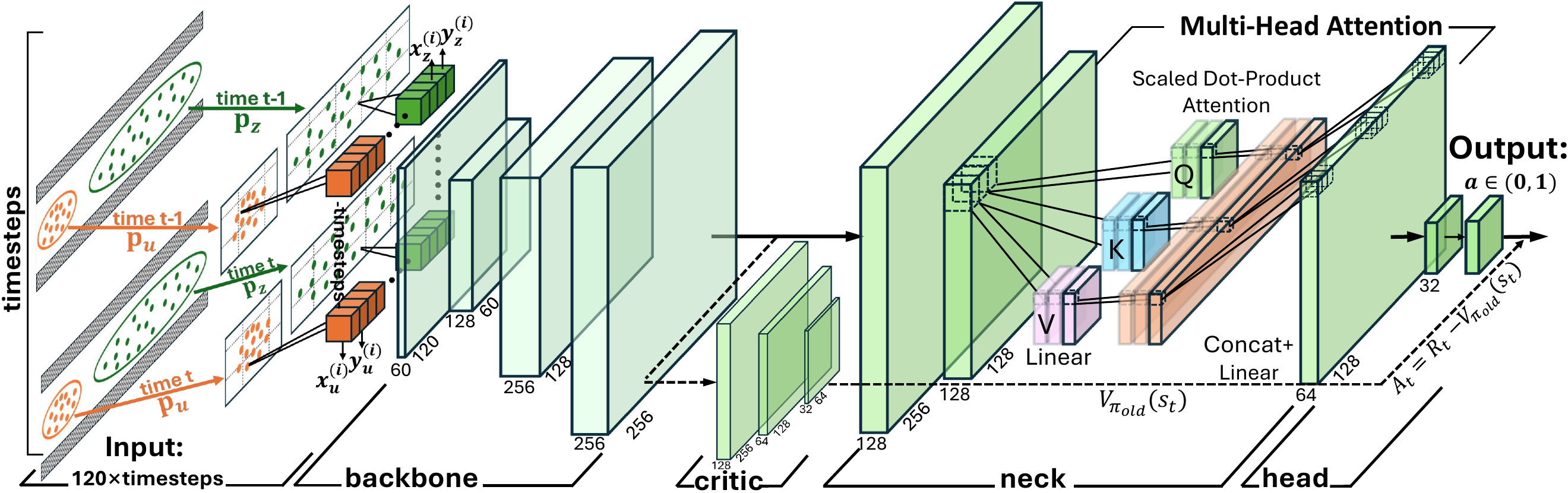}}

\caption{The architecture diagram of agent. The input part is divided into upper and lower parts, representing the data of two time frames respectively. Different colors represent $\textbf{p}_u$ and $\textbf{p}_z$ respectively.}

\label{agent_archi}
\end{figure*}

\section{PROBLEM FORMULATION}

Autonomously adjusting the contribution weights of different sensors is crucial for multi-sensor fusion. However, fixed-rule-based fusion lack flexibility and exhibit poor robustness. Meanwhile, traditional supervised learning methods, which heavily rely on the quality of dataset annotations, fail to accurately capture subtle degeneracy features. To address these issues, we use PPO to guide agent to dynamically change compensation strength of different sensors for pose optimization by autonomously adjusting degeneracy factor $a$.

Dynamic sensor compensation problem possesses the core characteristics of Markov Decision Process (MDP). We model this task using MDP, represented by tuple $\textbf{M}=(S, A, p, r, \gamma)$. SLAM is regarded as online interactive environment.
Furthermore, when we used (x, y, $\theta$) together as the input, we found that training effect is not good. This is because of the different pattern relationships between (x, y) and $\theta$.
Since robot mostly performs translational motions, $\sigma_{x}^{2}$ and $\sigma_{y}^{2}$ are more affected, so we focus on translational components $(x,y)$ of pose. Therefore, state $s \in \textbf{S}$ encapsulates pose $(x_u^{(i)},y_u^{(i)})$ in $\textbf{p}_u$ and pose $(x_z^{(i)},y_z^{(i)})$ in $\textbf{p}_z$, both of which are coordinates in the world frame.
Action $a \in \textbf{A}$ is degeneracy factor $a \in (0,1)$, which is published to SLAM via robot operating system (ROS). SLAM fine-tunes $\textbf{p}_z$ towards $\textbf{p}_u$ in proportion to $a$, to obtain fused $\textbf{p}_c$ integrating constraints from different sensors, and the best of $\textbf{p}_z$ and $\textbf{p}_c$ is chosen as the optimal pose according to likelihood score. Then SLAM updates and determine transition probability $p$, defining the likelihood of moving from state $s_i$ to state $s_j$ given action $a_i$. 
% However, since PPO is a model-free method, it does not directly estimate the transition probability but learns the policy through sampling and experience replay. Reward $r(s,a)$ will be discounted by $\gamma$. 

As the policy network in PPO, agent continuously adjusts policy $\pi^*$, and adapts action $a$ to dynamically compensate for degenerate state across different state spaces, so as to increase localization accuracy.
The policy $\pi^*$ is parameterized as $\theta$ by neural network. The input of policy network is state $s$, and output is probability distribution of action $a$. The goal is to find optimal parameter $\theta^*$ so that the policy $\pi^*$ in (\ref{pi}) can maximize the cumulative reward and learn the optimal action distribution in the abstract function space. 
\begin{equation}
\label{pi}
\pi^*=argmax_\pi V_\pi(s_0)
\end{equation}
PPO updates the policy parameter $\theta$ by optimizing the following objective function:
\begin{equation}
\label{L}
\begin{split}
L(\theta) &= \mathbb{E}_{s_t,a_t} [ \min ( \frac{\pi_{\theta}(a_t \vert s_t)}{\pi_{old}(a_t \vert s_t)} A_{\pi}(s_t,a_t),\\
&clip ( \frac {\pi_{\theta}(a_t \vert s_t)} {\pi_{old}(a_t \vert s_t)}, 1-\epsilon, 1+\epsilon ) A_{\pi} (s_{t},a_{t}) ) ]
\end{split}
\end{equation}
where $\pi_{old}$ is old strategy, $\epsilon$ is clipping parameter, $A_{\pi}(s_t,a_t)$ is advantage function, measuring the relative superiority or inferiority of action $a_t$ in state $s_t$.

The critic estimates value $V_{\pi}(s)$ in (\ref{V}), and it provides a baseline for agent to evaluate the quality of action, where $\textbf{$\tau$}$ is trajectory sequence $\{s_0,a_0,s_1,a_1,...,s_t,a_t\}$.
\begin{equation}
\label{V}
 V _ { \pi } ( s ) = E _ { \tau \sim \pi } \left[ \sum _ { t = 0 } ^ { \infty } \gamma ^ { t } r(s,a) \vert  s _ { 0 } = s \right]
\end{equation}

\section{Deep Reinforcement Learning}

In this section, we will explain the network architecture and specially designed reward.

\subsection{Degeneracy Optimization Agent}

The policy network serves as the critical decision-making part within agent, we process the output of SLAM into a vector
$\textbf{X}=\left[x_{z}^{(i)},y_{z}^{(i)},x_{u}^{(i)},y_{u}^{(i)}\vert ,i\in \left\{1,2,\ldots,n\right\}\right]$
as the input to agent, where $(x_z^{(i)},y_z^{(i)})$ represents the observation position of $i_{th}$ particle, and $(x_u^{(i)},y_u^{(i)})$ represents the position of $i_{th}$ particle given by motion model $u_m$ (specifically we set the number of particles to 30). We first analyze the coordinates as a whole quantity through a special weight matrix $\textbf{M}\in \mathbb{R}^{120 \times 60}$ (take (x, y) as input instead of taking x and y separately as input), so that network can learn the changes in distribution characteristics of particles in different environments.
The input passes through four linear layers with ReLU, reducing dimensions from 120 to 60, expanding to 128, and finally to 256. This process extracts higher-level features and enhances representation for downstream tasks. 
In the neck section, features are processed through two linear layers with ReLU, compressing dimensions from 256 to 128. They then pass through a Multi-Head Attention layer with 3 heads (each 64-dimensional) to capture complex relationships. This allows agent to learn the global relationships within particle swarm, and to capture the subtle signs of degeneracy according to particle divergence and Mahalanobis distance between $\textbf{p}_z$ and $\textbf{p}_u$, so as to improve the perception ability. 
The processed features are fed into a task-specific head which consists of a 64$\times$32 linear layer with ReLU and a 32$\times$1 linear layer with Sigmoid, output a degeneracy factor $a \in (0,1)$ to reflect the degree of degeneracy.
The critic network shares the backbone with policy to retain learned features. The head of critic uses three linear layers with ReLU to reduce features from 256 to 128, 64, 32, and outputs a value. The predicted values are used to compute gradients for updating policy network, guiding the update direction of policy.

\begin{figure*}[t]
	
	\begin{minipage}{0.245\linewidth}

		\centerline{\includegraphics[width=\textwidth]{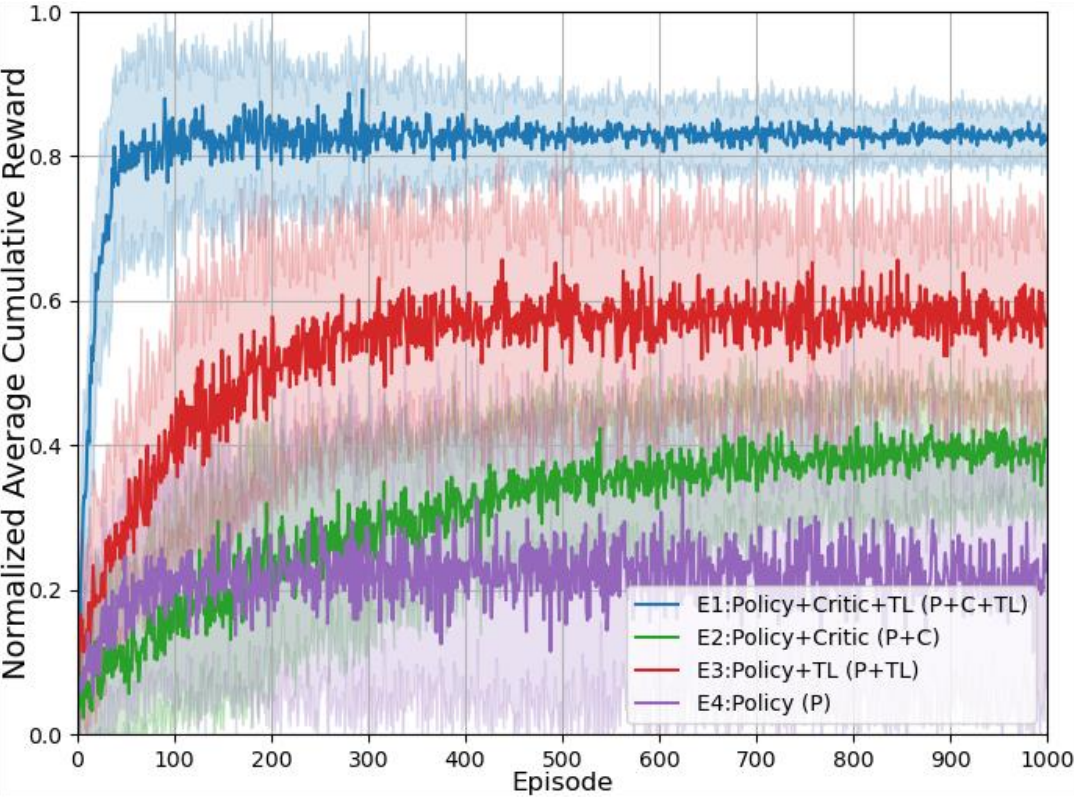}}
        \subcaption{E1-E4}

		% \centerline{Image}
	\end{minipage}
	\begin{minipage}{0.245\linewidth}
		
		\centerline{\includegraphics[width=\textwidth]{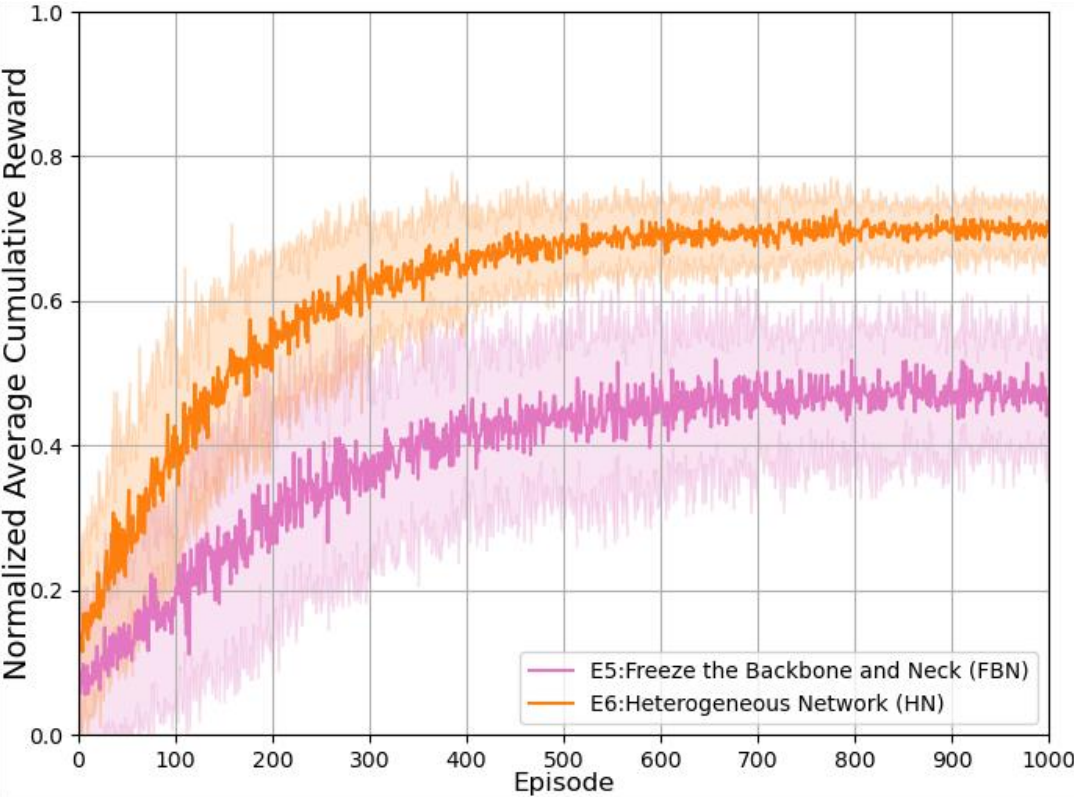}}
        \subcaption{E5-E6}
        
		% \centerline{Image}
	\end{minipage}
    \begin{minipage}{0.245\linewidth}
		
		\centerline{\includegraphics[width=\textwidth]{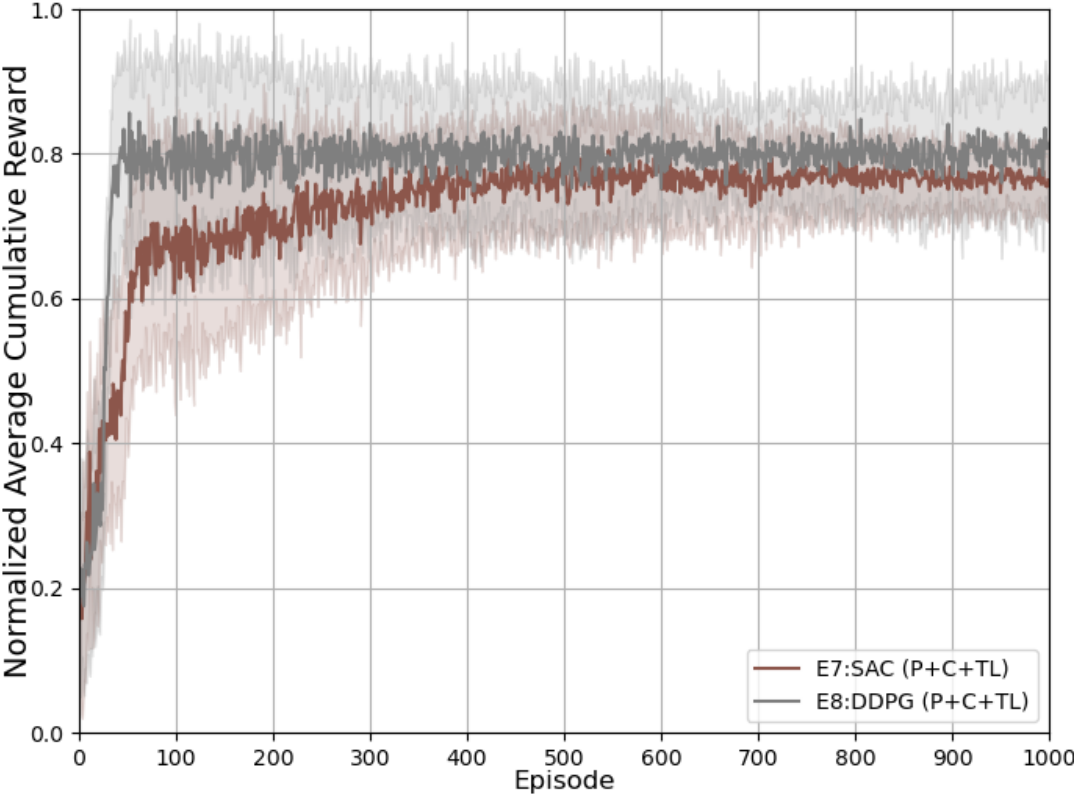}}
        \subcaption{E7-E8}
        
		% \centerline{Image}
	\end{minipage}
    \begin{minipage}{0.243\linewidth}
		
		\centerline{\includegraphics[width=\textwidth]{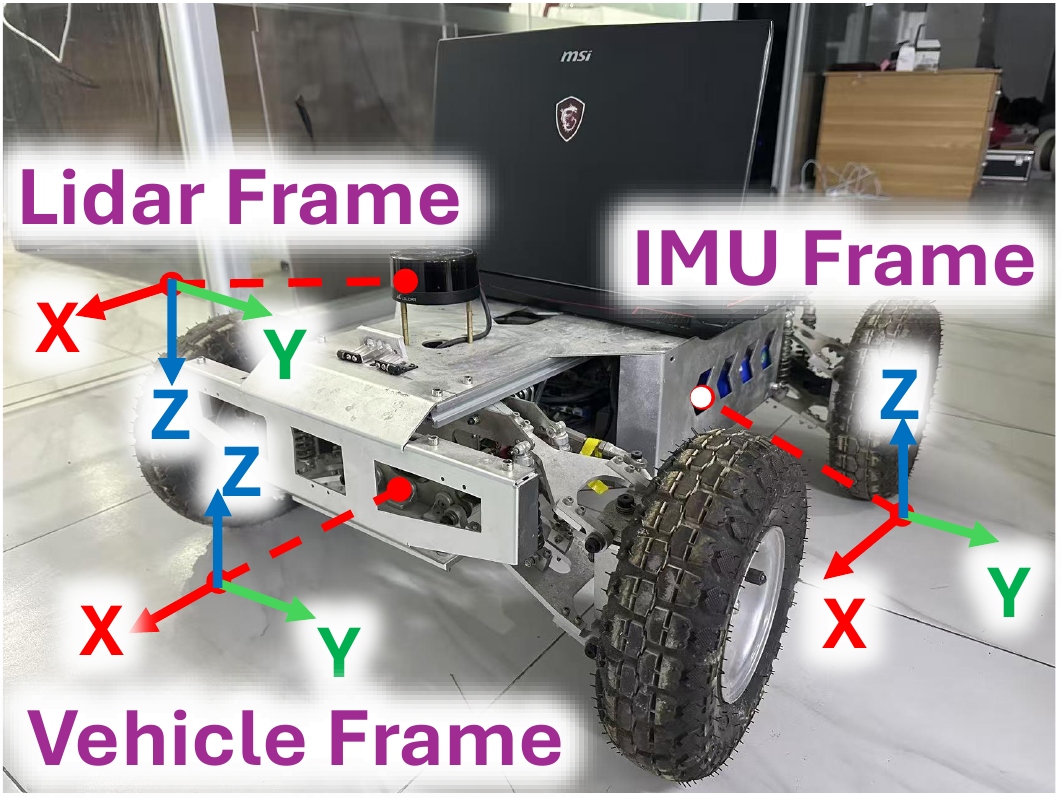}}
		% \centerline{Image}
        \subcaption{Robot and laptop}
        \label{instrument}
	\end{minipage}
    
	\caption{Reward curve of training process and experimental equipments (blue=E1=policy+critic+TL, green=E2=policy+critic, red=E3=policy+TL, purple=E4=policy, pink=E5=Freeze the backbone and neck, orange=E6=Heterogeneous network, grey=E7=SAC, brown=E8=DDPG).}
   
	\label{tl_result}
\end{figure*}

\subsection{Reward}

Reward is crucial for guiding agent to learn the optimal policy $\pi^*$. We use reward in (\ref{reward}) to prompt agent to output a rational degeneracy factor.
\begin{equation}
\label{reward}
\begin{split}
R &= \beta_{1} \cdot \exp\left(-\frac{1}{3}(\sigma_{x,t}^{2} + \sigma_{y,t}^{2})\right) + \beta_{2} \cdot s_t^{(i)} \\
&\quad + \beta_{3} \cdot N_{eff,t} - \beta_{4} \cdot \left\| a_{t} - a_{t-1} \right\|^{2}
\end{split}
\end{equation}
In (\ref{reward}), we use exponential decay term to encourage covariance ($\sigma_x^2$,$\sigma_y^2$) contraction, when agent effectively adjusts the constraint compensation, ($\sigma_x^2$,$\sigma_y^2$) will reduce, indicating improved localization accuracy. The reward increases rapidly to reinforce high-precision pose estimation when $(\sigma_x^2 + \sigma_y^2)$ is small.
And if optimized fused distribution $\textbf{p}_c$ achieves a better fit to obstacles in the map compared to original $\textbf{p}_z$, both localization accuracy and $s^{(i)}$ will increase.
Furthermore, particles carry weights in (\ref{weight}) that reflect the confidence of pose (important property carried by particles, but when used together with coordinate values as the input of agent, the training effect is not good). $N_{eff}$ represents the number of effective particles. A higher $N_{eff}$ indicates that the localization information of current particle is more accurate.
\begin{equation}
\label{weight}
w^{(i)}=p(\textbf{z}_{t}\vert \textbf{x}_{t},m) \cdot p(\textbf{x}_{t}\vert \textbf{x}_{t-1},\textbf{u}_{t})
\end{equation}
\begin{equation}
\label{N_{eff}}
N_{eff} = \frac {1} {\sum_{i=1}^{N}\overline{w}^{(i)}\cdot \overline{w}^{(i)}}
\end{equation}
where $\overline{w}^{(i)}$ represents the normalized weight of $i_{th}$ particle.

In addition, $\vert \vert a_{t}-a_{t-1}\vert \vert ^{2}$ constrains the difference between consecutive degeneracy factors, suppressing policy oscillation and ensuring stability of policy.
These metrics are normalized and weighted by $(\beta_1,\beta_2,\beta_3,\beta_4)$. The empirically best combination is $(\beta_1=0.3,\beta_2=0.3,\beta_3=0.2,\beta_4=0.2)$, derived through multiple training iterations.

\begin{table}[t]
\centering
\caption{Average rollout runtime of 1000 episodes.}

\label{runtime}
\renewcommand{\arraystretch}{0.8} % 调整行间距
\small % 直接调整整个表格的字体大小
\begin{tabular}{cc}
\hline
Components & Time(ms)  \\ \hline
Update frequency of SLAM & 170   \\ 
Network forward pass & 15.2 \\
ROS communication time & 4.3 \\ \hline
\end{tabular}

\end{table}

\section{Training \& Transfer Learning}

\subsection{Training Settings}

\begin{figure}[t]
\centerline{\includegraphics[width=0.5\textwidth]{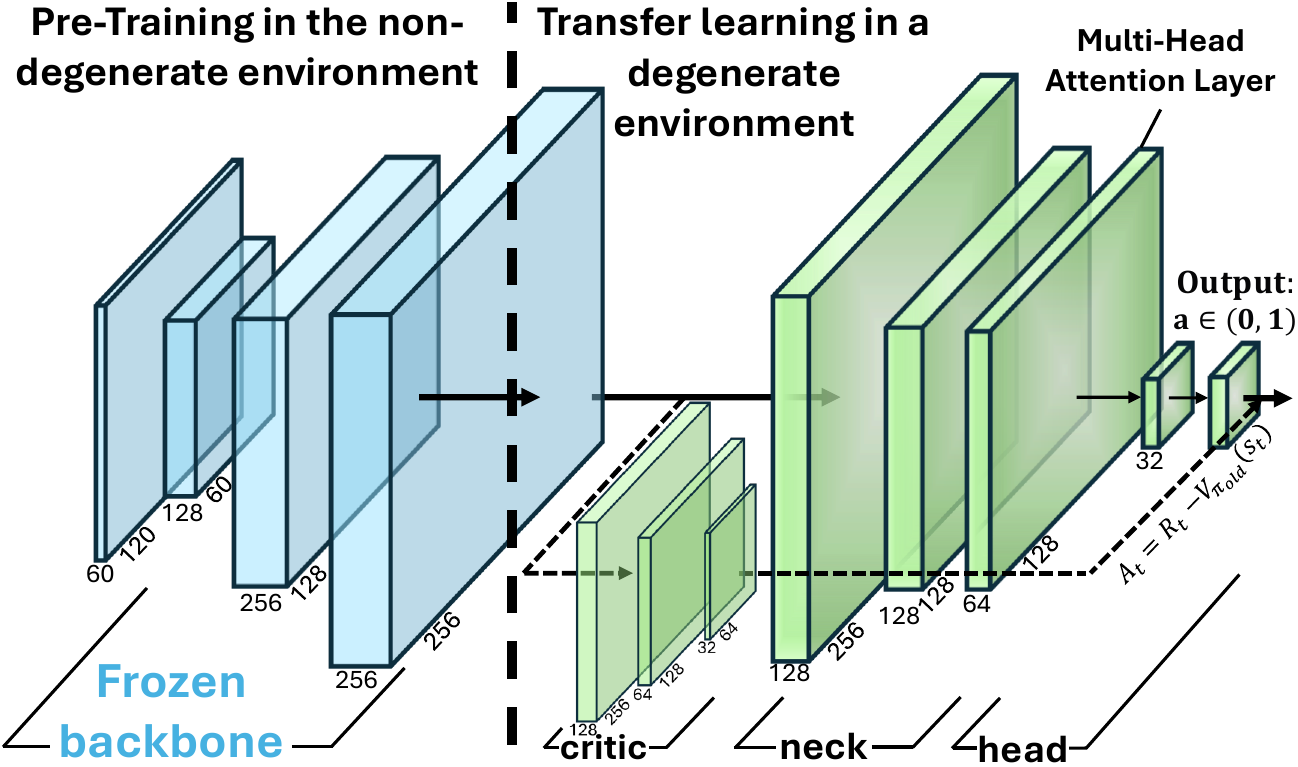}}

\caption{Schematic diagram of transfer learning.}

\label{transfer_learning_archi}
\end{figure}

\begin{table}[t]
\centering
\caption{Training hyperparameters.}

\label{parameter}
\renewcommand{\arraystretch}{0.8} 
\small 
\begin{tabular}{cccc}
\hline
Parameter & Value & Parameter & Value \\ \hline
Learning Rate & 0.002  & Discount Factor & 0.98  \\ 
Clip Factor & 0.2    & $T_{epochs}$     & 5  \\ 
Optimizer & Adam  & Timesteps  & 200   \\ 
Speed of Robot & 0.3m/s  & Episodes  & 1000 \\ \hline
\end{tabular}

\end{table}

\begin{figure*}[t]
	\centering
	\begin{subfigure}{0.21\linewidth}
		\centering
		\includegraphics[width=0.9\linewidth]{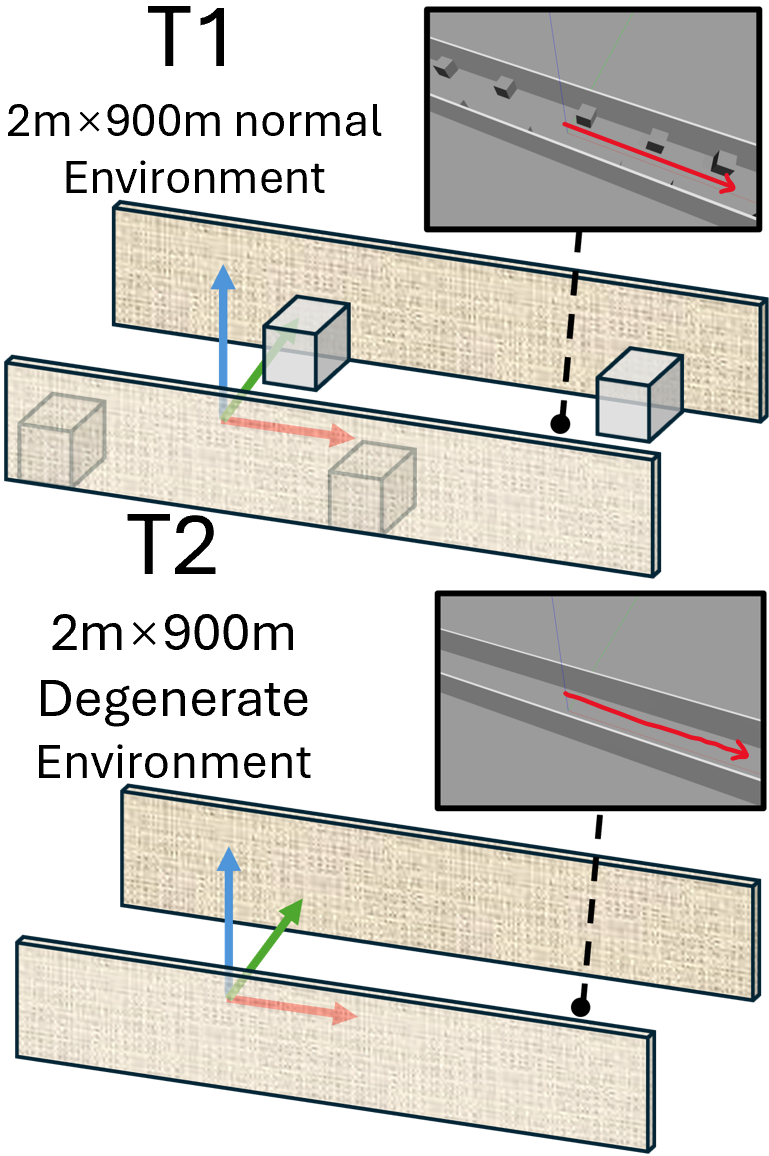}
		\caption{Training environments.}
		\label{env_train}%文中引用该图片代号
	\end{subfigure}
	\centering
	\begin{subfigure}{0.45\linewidth}
		\centering
		\includegraphics[width=0.9\linewidth]{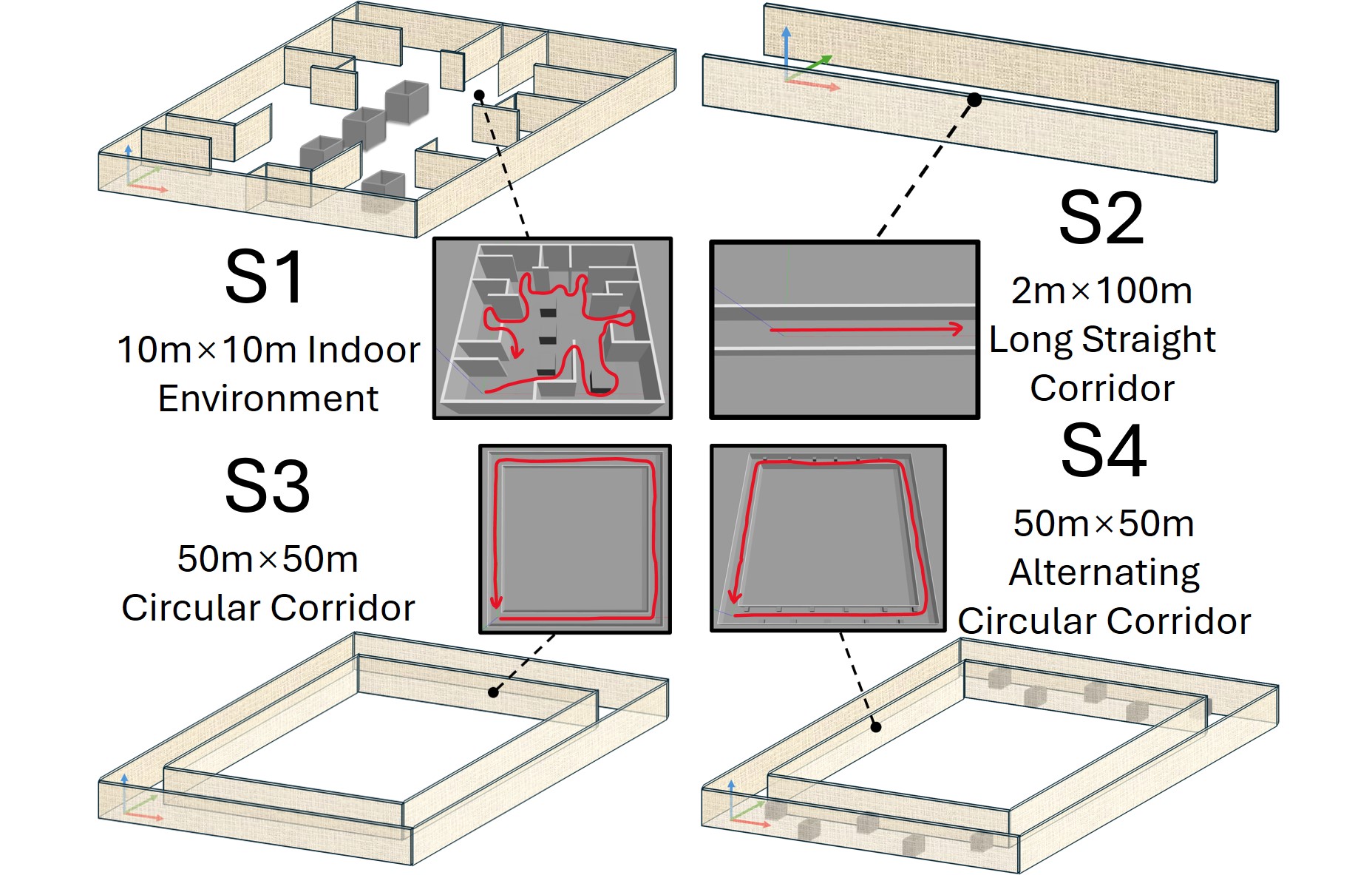}
		\caption{Simulation environments for evaluation.}
		\label{env_sim}%文中引用该图片代号
	\end{subfigure}
	\centering
	\begin{subfigure}{0.28\linewidth}
		\centering
		\includegraphics[width=0.9\linewidth]{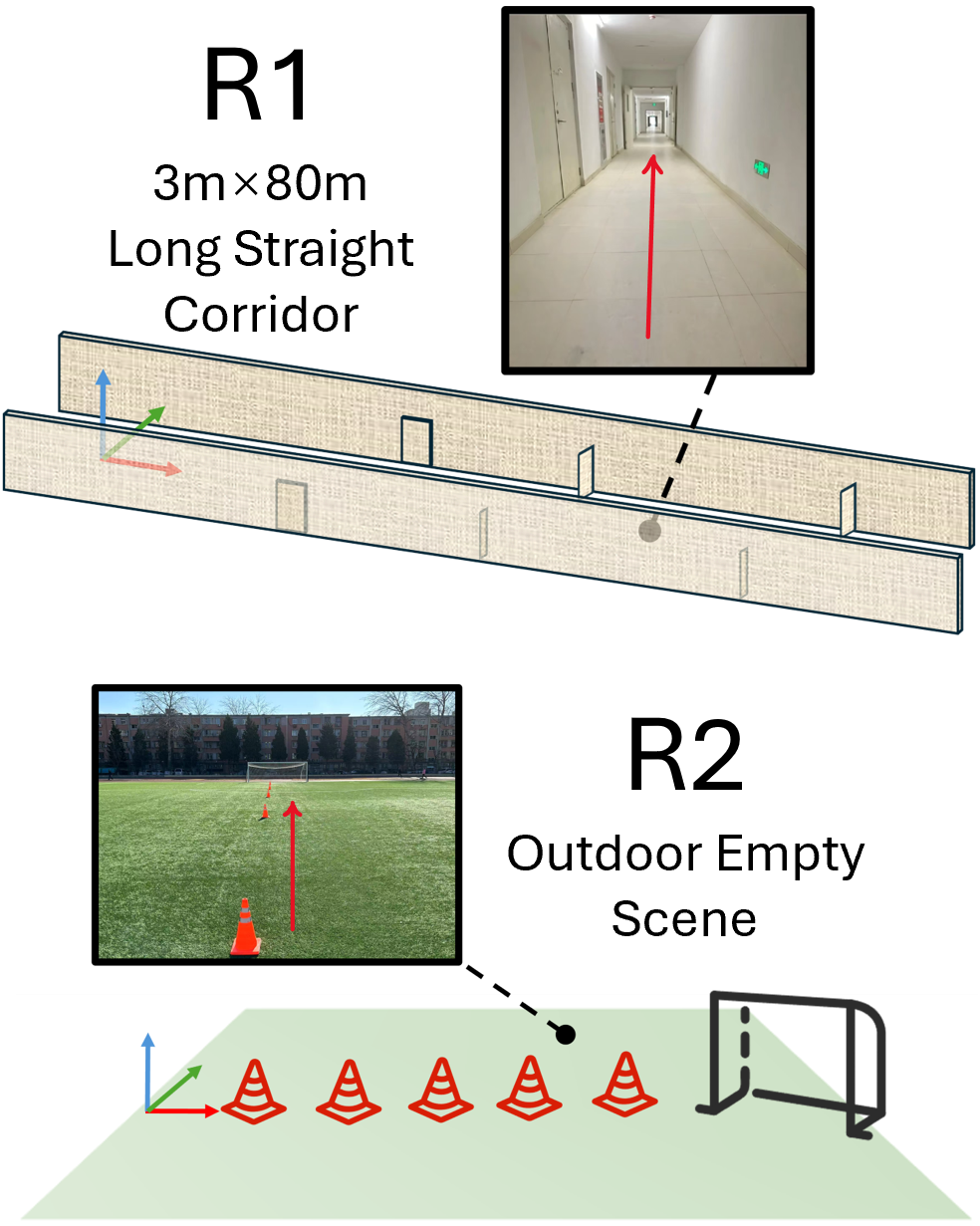}
		\caption{Real environments for evaluation.}
		\label{env_real}%文中引用该图片代号
	\end{subfigure}
   
	\caption{Diagram of the experimental environment with red motion trajectories.}
   
	\label{env}
\end{figure*}

About the training platform, we adopt a computer with the ubuntu 20.04 and GMapping \cite{grisetti2007improved}, equipped with a 13th Gen Intel Core i7-13700K CPU and an NVIDIA GeForce RTX 2060 Super GPU.
Before training begins, we set all hyperparameters according to Table. \ref{parameter}, which are empirically best combinations derived from multiple experiments. Since the architecture of agent is difficult to optimize through selection, its rationality is validated through experimental verification.
We aim for the agent to be robust in various scenarios. The experimental scenes are divided into non-degenerate (with rich features and obstacles) and degenerate (long corridors with sparse features), as shown in Fig. \ref{env_train}.

 \begin{figure}[t]
\centerline{\includegraphics[width=0.5\textwidth]{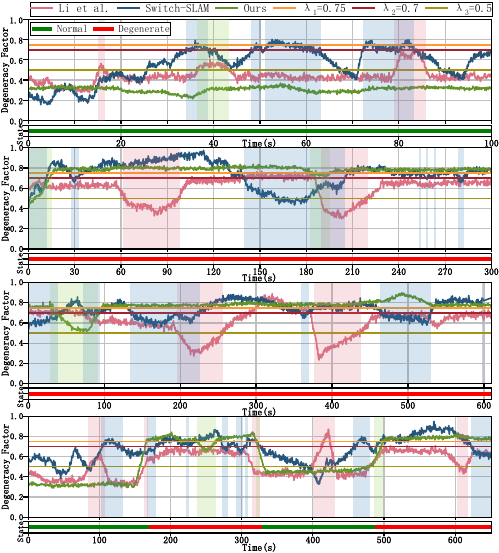}}

\caption{Simulation results of degeneracy detection.}

\label{dd_sim}
\end{figure}

\begin{figure}[t]
\centerline{\includegraphics[width=0.5\textwidth]{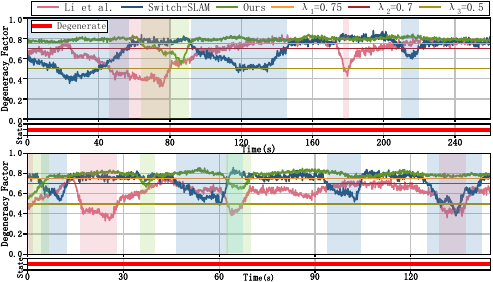}}

\caption{Real experimental results of degeneracy detection.}

\label{dd_real}
\end{figure}

\subsection{Training Details \& Results}

Training the agent directly in degenerate scene is challenging due to the sparsity of features.
Instead, we first train the agent in a non-degenerate scene. After initial training, we use transfer learning to retain the decision-making capabilities of agent from non-degenerate scene and apply them to degenerate scene. Specifically, we freeze the backbone of policy and critic networks and fine-tune the remaining layers in degenerate scene, update parameters for the rest of policy and critic, as shown in Fig. \ref{transfer_learning_archi}. 
It should be noted that transfer learning does not leak the privileged information of the environment into the policy network, because essentially SLAM optimizes its own model distribution to maximize the reward function and capture representative signs of degeneracy.

We conduct eight sets of experiments (E1-E8), each with five training epochs, recording and normalizing the cumulative rewards per episode. The results are shown in Fig. \ref{tl_result}, with lines representing mean values and shaded areas indicating standard deviations.
The training process takes 14 hours in E1, the agent is trained in a non-degenerate scene and transferred to a degenerate scene using transfer learning (TL). It quickly adapt, achieving stable reward of around 0.82, with 35\% increase in peak reward compared to E2. 
In contrast, agent train directly in the degenerate scene (E2) show lower performance.
Removing the critic module (E3) lead to a 0.2 drop in reward and higher variance compared to E1, highlighting the role of critic in decision-making. Training the policy directly in degenerate scene (E4) result in poor performance and a reward of around 0.2, further emphasizing the importance of TL and the critic for effective decision-making. In E5, we freeze the backbone and neck, updating only the head network and critic, the reward dropped by 0.3 compared to E1, this indicates that limited training flexibility reduces the decision-making performance of agent. In E6, we separate the critic into an independent network. The reward decrease by 0.15, showing that critic benefits from shared backbone. And we present the runtime of main components in Table. \ref{runtime}.

Additionally, we compare PPO with DDPG \cite{lillicrap2019continuouscontroldeepreinforcement} and SAC \cite{christodoulou2019softactorcriticdiscreteaction}. The target and main networks of DDPG match the architecture of our method, while the dual-critic architecture of SAC shares bottom network with actor. We freeze main network of DDPG and entire network of SAC, training them in both non-degenerate and degenerate scenes. 
The reward of SAC (E7) stabilize around 0.77 with low fluctuations, whereas the reward of DDPG (E8) reach higher peaks around 0.8 but with larger variations. PPO achieve the highest reward and best performance. Next, we will validate the performance and real-time computational efficiency of PPO.

\begin{table}[t]
    \caption{Results of degeneracy detection, the underlined numbers represent the best performance of each method.}
   
    \label{dd_table}
    \centering
    \resizebox{1\columnwidth}{!}{
    \begin{tabular}{c|c cccccc}
      \toprule[0.5mm]
      \textbf{Scene} & \textbf{Method} & \textbf{S1} & \textbf{S2} & \textbf{S3} & \textbf{S4} & \textbf{R1} & \textbf{R2}\\
      
      \midrule
      \midrule
      \multirow{13}{*}{\rotatebox{90}{\textit{\textbf{Success Ratio(\%)}}}}
      % \multirow{12}{*}{\rotatebox{90}{\textit{Ratio(\%)}}}
      &&& \textbf{Baseline}\\
      \cmidrule{2-8}
      &Switch-SLAM \cite{lee2024switch} & 73.6  & 72.3 & 64.7 & \underline{84.5} & 55.4 & 65.9  \\ 
      &Li $\textit{et al.}$ \cite{10994466} & 85.3  & 78.3 & 81.2 & 83.7 & \underline{85.9} & 82  \\

      \cmidrule{2-8}
      &&& \textbf{Ablation}\\
      \cmidrule{2-8}
      &Ours (P+C) & \underline{51.6} & 48.3 & 46.3  
      & 42.2 & 42.1 & 39.6 \\
      
      &Ours (P+TL) & \underline{72.9} & 69.5 & 65.1 
      & 64.5 & 62.1 & 67.2 \\
      
      &Ours (P) & \underline{35.1} & 30.8 & 31.2 
      & 33.2 & 30.1 & 34.2 \\

      &Ours (FBN) & 46.3 & \underline{46.5} & 43.2 
      & 44.9 & 42.7 & 44.9 \\

      &Ours (HN) & 79.6 & 75.3 & 78.5
      & 77.9 & \underline{81.2} & 80.4 \\
      
      \cmidrule{2-8}
      &Ours (P+C+TL) & \underline{\textbf{100}} & \underline{\textbf{90.5}} & \underline{\textbf{89.2}}  
      &\underline{\textbf{91}} & \underline{\textbf{89.6}} & \underline{\textbf{87.3}} \\
      
      \midrule
      \midrule
      \multirow{4}{*}{\rotatebox{90}{\textit{\textbf{Inference}}}}
      \multirow{4}{*}{\rotatebox{90}{\textit{\textbf{Time(ms)}}}}
      &Switch-SLAM \cite{lee2024switch} & 15.2 & \underline{\textbf{9.5}} & 12.4 & 12.2 & 15.3 & 15.8  \\
      &Li $\textit{et al.}$ \cite{10994466} & \underline{\textbf{10.5}}  & 12.2 & 11.8 & 13.7 & 20.4 & 21.2  \\
      
      \cmidrule{2-8}
      & Ours & 15.6 & 15.8 & \underline{\textbf{14.7}}  & 15.2 & 18.5 & 17.9\\
      \bottomrule[0.5mm]
      
    \end{tabular}}
   
\end{table}

\begin{table*}[t]
\centering
\caption{Results of degeneracy optimization (Simulations).}

\label{do_result_sim}
\resizebox{\textwidth}{!}{%
\begin{tabular}{|c|c|c|ccc|ccc|ccc|ccc|}
\hline
\multirow{3}{*}{Method} & \multirow{3}{*}{Odometry} & \multirow{3}{*}{\begin{tabular}[c]{@{}c@{}}Average\\ Speed\end{tabular}} & \multicolumn{3}{c|}{S1} & \multicolumn{3}{c|}{S2} & \multicolumn{3}{c|}{S3} & \multicolumn{3}{c|}{S4} \\ \cline{4-15} 
 &  &  & \multicolumn{1}{c|}{\multirow{2}{*}{ATE}} & \multicolumn{2}{c|}{MAX} & \multicolumn{1}{c|}{\multirow{2}{*}{ATE}} & \multicolumn{2}{c|}{MAX} & \multicolumn{1}{c|}{\multirow{2}{*}{ATE}} & \multicolumn{2}{c|}{MAX} & \multicolumn{1}{c|}{\multirow{2}{*}{ATE}} & \multicolumn{2}{c|}{MAX} \\ \cline{5-6} \cline{8-9} \cline{11-12} \cline{14-15} 
 &  &  & \multicolumn{1}{c|}{} & \multicolumn{1}{c|}{x} & y & \multicolumn{1}{c|}{} & \multicolumn{1}{c|}{x} & y & \multicolumn{1}{c|}{} & \multicolumn{1}{c|}{x} & y & \multicolumn{1}{c|}{} & \multicolumn{1}{c|}{x} & \multicolumn{1}{c|}{y} \\ \hline

Cartographer \cite{7487258} &\cellcolor[HTML]{CBCEFB}\ding{55}  &0.286 m/s  
& \multicolumn{1}{c|}{0.112} & \multicolumn{1}{c|}{0.153} 
& \multicolumn{1}{c|}{0.149}  & \multicolumn{1}{c|}{-} 
&\multicolumn{1}{c|}{-} & \multicolumn{1}{c|}{-} 
&\multicolumn{1}{c|}{-}  & \multicolumn{1}{c|}{-} 
& \multicolumn{1}{c|}{-} &\multicolumn{1}{c|}{-}  
& \multicolumn{1}{c|}{-} & \multicolumn{1}{c|}{-}  \\ \hline
 
Hector SLAM \cite{kohlbrecher2011flexible} &\cellcolor[HTML]{CBCEFB}\ding{55}  &0.289 m/s  
& \multicolumn{1}{c|}{0.287} & \multicolumn{1}{c|}{0.295} 
& \multicolumn{1}{c|}{0.307}  & \multicolumn{1}{c|}{4.411} 
&\multicolumn{1}{c|}{4.97} & \multicolumn{1}{c|}{4.212} 
&\multicolumn{1}{c|}{7.752}  & \multicolumn{1}{c|}{8.355} 
& \multicolumn{1}{c|}{8.613} &\multicolumn{1}{c|}{4.657}  
& \multicolumn{1}{c|}{5.531} & \multicolumn{1}{c|}{5.416}  \\ \hline

Li $\textit{et al.}$ \cite{10994466}   &\cellcolor[HTML]{FFCCC9}$\checkmark$  &0.284 m/s  
& \multicolumn{1}{c|}{0.126} & \multicolumn{1}{c|}{0.185} 
& \multicolumn{1}{c|}{0.157}  & \multicolumn{1}{c|}{0.196} 
&\multicolumn{1}{c|}{0.251} & \multicolumn{1}{c|}{0.357} 
&\multicolumn{1}{c|}{0.392}  & \multicolumn{1}{c|}{0.578} 
& \multicolumn{1}{c|}{0.785} &\multicolumn{1}{c|}{0.285}  
& \multicolumn{1}{c|}{0.35} & \multicolumn{1}{c|}{0.411}  \\ \hline

GMapping \cite{grisetti2007improved} &\cellcolor[HTML]{FFCCC9}$\checkmark$  &0.291 m/s  &\multicolumn{1}{c|}{0.159} & \multicolumn{1}{c|}{0.188} 
& \multicolumn{1}{c|}{0.142}  & \multicolumn{1}{c|}{0.692} 
&\multicolumn{1}{c|}{0.983} & \multicolumn{1}{c|}{0.861} 
&\multicolumn{1}{c|}{4.32}  & \multicolumn{1}{c|}{5.124} 
& \multicolumn{1}{c|}{4.675} &\multicolumn{1}{c|}{2.561}  
& \multicolumn{1}{c|}{3.412} & \multicolumn{1}{c|}{2.869}  \\ \hline

Ours (GMapping+DOA) &\cellcolor[HTML]{FFCCC9}$\checkmark$  &0.295 m/s  
&\multicolumn{1}{c|}{0.102} & \multicolumn{1}{c|}{0.124} 
& \multicolumn{1}{c|}{0.137}  & \multicolumn{1}{c|}{0.126} 
&\multicolumn{1}{c|}{0.254} & \multicolumn{1}{c|}{0.187} 
&\multicolumn{1}{c|}{0.266}  & \multicolumn{1}{c|}{0.358} 
& \multicolumn{1}{c|}{0.332} &\multicolumn{1}{c|}{0.122}  
& \multicolumn{1}{c|}{0.186} & \multicolumn{1}{c|}{0.136}  \\ \hline
\end{tabular}%
}
\end{table*}

\begin{table}[t]
\centering

\caption{Results of degeneracy optimization (Real-World).}

\label{do_real_resample}
\resizebox{\columnwidth}{!}{%
\begin{tabular}{|c|c|c|cc|cc|}
\hline
\multirow{3}{*}{Method} &
  \multirow{3}{*}{Odometry} &
  \multirow{3}{*}{\begin{tabular}[c]{@{}c@{}}Average\\ Speed\end{tabular}} &
  \multicolumn{2}{c|}{\multirow{2}{*}{\begin{tabular}[c]{@{}c@{}}Resample\\ Frequency\end{tabular}}} &
  \multicolumn{2}{c|}{\multirow{2}{*}{$N_{eff}$}} \\
&  &  & \multicolumn{2}{c|}{}        & \multicolumn{2}{c|}{}        \\ \cline{4-7} 
&  &  & \multicolumn{1}{c|}{R1} & R2 & \multicolumn{1}{c|}{R1} & R2 \\ \hline

Cartographer \cite{7487258}  & \cellcolor[HTML]{CBCEFB}\ding{55}  & 0.248 m/s & \multicolumn{1}{c|}{-}   & -   
&\multicolumn{1}{c|}{-}   &  -  \\ \hline

Hector SLAM \cite{kohlbrecher2011flexible}  & \cellcolor[HTML]{CBCEFB}\ding{55}  & 0.257 m/s & \multicolumn{1}{c|}{-}   & -   
&\multicolumn{1}{c|}{-}   &  -  \\ \hline

Li $\textit{et al.}$ \cite{10994466} & \cellcolor[HTML]{FFCCC9}$\checkmark$ & 0.273 m/s & \multicolumn{1}{c|}{47} & 49 & \multicolumn{1}{c|}{20.1} & 19.5 \\ \hline

GMapping \cite{grisetti2007improved} & \cellcolor[HTML]{FFCCC9}$\checkmark$  & 0.26 m/s & \multicolumn{1}{c|}{44} & 52 & \multicolumn{1}{c|}{18.6} & 16.2   \\ \hline

Ours (GMapping+DOA) & \cellcolor[HTML]{FFCCC9}$\checkmark$ & 0.276 m/s & \multicolumn{1}{c|}{26} & 33 & \multicolumn{1}{c|}{22.3} & 21.3 \\ \hline
\end{tabular}%
}
\end{table}

\section{EXPERIMENTAL EVALUATION}

\subsection{Experiment Settings} 

We use a laptop computer with an Intel i7-7700HQ CPU, an NVIDIA GeForce GTX 1060 GPU and 16GB RAM to remotely maneuver an wheeled robot. 
The robot is equipped with an LSlidar M10P lidar, which is powered by an ARM Cortex-A72 64-bit CPU, a Broadcom VideaCore VI GPU and 4GB RAM, as shown in Fig. \ref{tl_result}. The simulation platform is the same as the agent training platform.
We set up four simulation environments Fig. \ref{env_sim} S1-S4 and two real environments Fig. \ref{env_real} R1-R2 as experiment scenes.

\subsection{Degeneracy Detection}

In degeneracy detection experiment, we load parameters from the peak reward moment during training and remove optimization module. Although the dynamic response of reward change, degeneracy factor still encodes the latent projection mapping of environmental features. The network parameters establish a stable, nonlinear correspondence between environmental degeneracy and hidden layer activations. This correlation persists even after removing optimization module.

We conduct degeneracy detection experiments in six scenes using GMapping (30 particles), with the robot moving at 0.3 m/s. We first extract the Hessian matrix from observation model to calculate the factor in \cite{lee2024switch}, normalize, invert, and correct the factor by adding 1. We compare our method with \cite{lee2024switch} and \cite{10994466}, recording detection results for each time frame.
The degeneracy factor curves are shown in Fig. \ref{dd_sim}, with shaded areas indicating incorrect judgment periods. We set our degeneracy threshold at 0.75, as agent outputs in degenerate scenes are mostly above this value. Method \cite{10994466} uses a threshold of 0.5, and \cite{lee2024switch} uses 0.7, with all factors normalized to follow the rule "larger factor means higher degeneracy."
We also conduct ablation experiments of the architecture, calculate the success rate and average inference time, as shown in Table \ref{dd_table}.

Our proposed method achieves detection success rate of around 90\%, with no false detections in simple indoor scenes and strong generalization in real-world scenes. In Fig. \ref{dd_real}a, our method improves detection accuracy by 34.2\% compared to Switch-SLAM. As shown in Fig. \ref{dd_sim}d, factor curve of our method changes rapidly at road junctions, indicating quick adaptation to environmental changes and guiding optimization adjustments. The curve remains smooth during other periods, showing the ability of agent to adjust sensor contributions smoothly.
In contrast, Switch-SLAM shows only a 0.8\% higher accuracy than \cite{10994466} in Fig. \ref{dd_real}a and performs poorly in other experiments, indicating limited detection capability. While \cite{10994466} maintains 80\% accuracy, it falls short compared to our method. Its steep factor curve suggests limited fine-tuning ability and inability to capture subtle environmental changes effectively. In ablation study, we test different architectures and found their detection accuracies are relatively low, which shows the rationality of our architecture design.

Additionally, Switch-SLAM has shorter inference time, our method has relatively longer inference time compared to the other methods but is still significantly lower than the 170ms frequency of SLAM. The experiments show that our method excels in detecting degeneracy across various scenes, accurately capturing environmental changes and providing precise results while maintaining real-time performance of SLAM.

\begin{figure}[t]
\centerline{\includegraphics[width=0.5\textwidth]{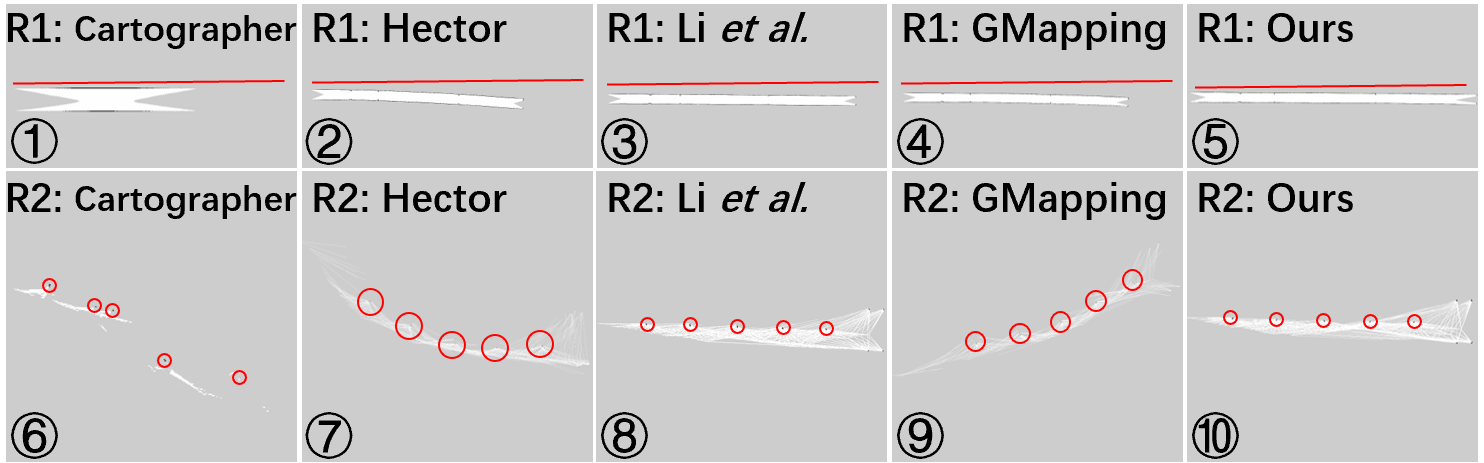}}

\caption{Maps of experiments. The red line representing the true length serves as a reference for the degree of map skewing and red circle is the location of landmark.}
\label{do_real_map}
\end{figure}

\subsection{Degeneracy Optimization}

\subsubsection{Simulations}

We demonstrate the performance of DOA through ablation experiments using GMapping \cite{grisetti2007improved} as baseline.
Experiments are conducted in four simulation environments (Fig. \ref{env_sim}) with Cartographer \cite{7487258}, Hector \cite{kohlbrecher2011flexible}, \cite{10994466}, original GMapping \cite{grisetti2007improved} and DOA-GMapping. We calculate absolute trajectory error of translation (ATE) using \cite{grupp2017evo}, the results are shown in Table. \ref{do_result_sim}, we also calculate the maximum values (MAX) in meters of the difference between estimated pose and GT provided by Gazebo on x and y axes.

In S1, all these five methods have good performance, and DOA reduces the ATE of GMapping by about 35.8\%, which is 0.024 less than \cite{10994466}, indicating that DOA can improve the localization of SLAM in indoor environment.
In S2, S3 and S4, DOA reduces the ATE of GMapping by 81.8\%, 93.8\% and 95.2\% respectively, indicating that the degeneracy of SLAM is significantly improved. And compared with \cite{10994466}, the accuracy in these scenarios is improved by 35.7\%, 32.1\% and 57.2\% respectively. Secondly, the maximum value of pose error of DOA on x and y axis is relatively small, indicating that DOA can accurately capture the degeneracy of environmental features, so as to timely compensate constraint information. Hector \cite{kohlbrecher2011flexible} has the worst performance due to lack of odometry information compensation. 
In addition, Cartographer \cite{7487258} encounter localization failure problems in all these three scenarios.
The above experiments prove that DOA can perform adaptive compensation for pose estimation in various environments to reduce the influence of degeneracy.

\subsubsection{Real Experiments}

We verify the performance of DOA in two real-world scenes (Fig. \ref{env_real}). Since GT of localization is difficult to obtain, we analyze qualitatively by map quality, as shown in Fig. \ref{do_real_map}. In addition, we record the number of resample in each experiment to evaluate the maintenance of particle diversity, as shown in Table. \ref{do_real_resample}.
In R1, the length of map constructed by our method is closest to the true length (red line) compared with other methods.
It can be seen that corridor map constructed by Cartographer has the problem of movement stagnation in R1. The length of map differs greatly from actual value, and the offset of landmarks in R2 is relatively large. And localization of Hector has large drift in degenerate scenes, the environment structures in maps appear to be skewed to a large extent. At the same time, maps constructed by GMapping also distort because there is no constraint compensation process. Although our method is similar to Li $\textit{et al.}$ \cite{10994466} in terms of the quality of map, the structure of corridor in R1 and relative positions of landmarks in R2 are accurate, it can be seen from quantitative indicators in Table. \ref{do_real_resample} that our method maintains particle diversity well, reduces the number of resample by adaptive pose compensation strategy and extends the life cycle of particles with high fitting degree.
Based on the performance of DOA in different scenes above, we shows that transfer learning can enhance the generalization ability of DOA.

\section{Conclusion}

This paper has presented an online simulation training framework based on PPO. The method has overcome the issue of ambiguous labeling rules in traditional supervised learning. By interacting with SLAM in real-time and using a specially designed reward, the framework guided agent to autonomously adjust the contribution of different sensors to pose optimization in degenerate scenes. Specifically, the agent fine-tuned observation distribution along the line connecting centroids of observation and motion model distributions. The degeneracy factor output by agent controled the distance of this adjustment towards motion model, thereby compensating for lost constraints.
Additionally, we employed transfer learning to enhance the generalization ability of agent across different scenes and improved its training performance in degenerate scenes. Through simulations and real-world experiments, we demonstrated the robust degeneracy detection and optimization capabilities of agent in various scenes.
In the future, we will use particle distribution moments or embeddings to enable the generalization of DOA to SLAM systems with varying particle counts.

% References

\bibliographystyle{Bibliography/IEEEtranTIE}
\bibliography{reference}\ %IEEEabrv instead of IEEEfull

\begin{thebibliography}{10}
\providecommand{\url}[1]{#1}
\csname url@samestyle\endcsname
\providecommand{\newblock}{\relax}
\providecommand{\bibinfo}[2]{#2}
\providecommand{\BIBentrySTDinterwordspacing}{\spaceskip=0pt\relax}
\providecommand{\BIBentryALTinterwordstretchfactor}{4}
\providecommand{\BIBentryALTinterwordspacing}{\spaceskip=\fontdimen2\font plus
\BIBentryALTinterwordstretchfactor\fontdimen3\font minus \fontdimen4\font\relax}
\providecommand{\BIBforeignlanguage}[2]{{%
\expandafter\ifx\csname l@#1\endcsname\relax
\typeout{** WARNING: IEEEtran.bst: No hyphenation pattern has been}%
\typeout{** loaded for the language `#1'. Using the pattern for}%
\typeout{** the default language instead.}%
\else
\language=\csname l@#1\endcsname
\fi
#2}}
\providecommand{\BIBdecl}{\relax}
\BIBdecl

\bibitem{schulman2017proximalpolicyoptimizationalgorithms}
\BIBentryALTinterwordspacing
J.~Schulman, ``Proximal policy optimization algorithms,'' 2017. [Online]. Available: \url{https://arxiv.org/abs/1707.06347}
\BIBentrySTDinterwordspacing

\bibitem{grisetti2007improved}
G.~Grisetti, ``Improved techniques for grid mapping with rao-blackwellized particle filters,'' \emph{IEEE transactions on Robotics}, vol.~23, no.~1, pp. 34--46, 2007.

\bibitem{7487258}
W.~Hess, ``Real-time loop closure in 2d lidar slam,'' in \emph{2016 IEEE International Conference on Robotics and Automation (ICRA)}, \href{http://dx.doi.org/10.1109/ICRA.2016.7487258}{DOI 10.1109/ICRA.2016.7487258}, pp. 1271--1278, 2016.

\bibitem{kohlbrecher2011flexible}
S.~Kohlbrecher, ``A flexible and scalable slam system with full 3d motion estimation,'' in \emph{2011 IEEE international symposium on safety, security, and rescue robotics}, pp. 155--160.\hskip 1em plus 0.5em minus 0.4em\relax IEEE, 2011.

\bibitem{campos2021orb}
C.~Campos, ``Orb-slam3: An accurate open-source library for visual, visual--inertial, and multimap slam,'' \emph{IEEE Transactions on Robotics}, vol.~37, no.~6, pp. 1874--1890, 2021.

\bibitem{10740921}
Z.~Chen, ``A fast and accurate visual inertial odometry using hybrid point-line features,'' \emph{IEEE Robotics and Automation Letters}, vol.~9, \href{http://dx.doi.org/10.1109/LRA.2024.3490406}{DOI 10.1109/LRA.2024.3490406}, no.~12, pp. 11\,345--11\,352, 2024.

\bibitem{10305271}
A.~Kumar, ``High-speed stereo visual slam for low-powered computing devices,'' \emph{IEEE Robotics and Automation Letters}, vol.~9, \href{http://dx.doi.org/10.1109/LRA.2023.3329621}{DOI 10.1109/LRA.2023.3329621}, no.~1, pp. 499--506, 2024.

\bibitem{10874215}
K.~Xu and S.~Yuan, ``Airslam: An efficient and illumination-robust point-line visual slam system,'' \emph{IEEE Transactions on Robotics}, \href{http://dx.doi.org/10.1109/TRO.2025.3539171}{DOI 10.1109/TRO.2025.3539171}, pp. 1--20, 2025.

\bibitem{zhang2014loam}
J.~Zhang, ``Loam: Lidar odometry and mapping in real-time.'' in \emph{Robotics: Science and systems}, vol.~2, no.~9, pp. 1--9.\hskip 1em plus 0.5em minus 0.4em\relax Berkeley, CA, 2014.

\bibitem{shan2018lego}
T.~Shan, ``Lego-loam: Lightweight and ground-optimized lidar odometry and mapping on variable terrain,'' in \emph{2018 IEEE/RSJ International Conference on Intelligent Robots and Systems (IROS)}, pp. 4758--4765.\hskip 1em plus 0.5em minus 0.4em\relax IEEE, 2018.

\bibitem{10757429}
C.~Zheng, ``Fast-livo2: Fast, direct lidar–inertial–visual odometry,'' \emph{IEEE Transactions on Robotics}, vol.~41, \href{http://dx.doi.org/10.1109/TRO.2024.3502198}{DOI 10.1109/TRO.2024.3502198}, pp. 326--346, 2025.

\bibitem{10404014}
T.~Wen, ``Liver: A tightly coupled lidar-inertial-visual state estimator with high robustness for underground environments,'' \emph{IEEE Robotics and Automation Letters}, vol.~9, \href{http://dx.doi.org/10.1109/LRA.2024.3355778}{DOI 10.1109/LRA.2024.3355778}, no.~3, pp. 2399--2406, 2024.

\bibitem{10776572}
J.~Xu, ``Intermittent vio-assisted lidar slam against degeneracy: Recognition and mitigation,'' \emph{IEEE Transactions on Instrumentation and Measurement}, vol.~74, \href{http://dx.doi.org/10.1109/TIM.2024.3507053}{DOI 10.1109/TIM.2024.3507053}, pp. 1--13, 2025.

\bibitem{10631284}
Q.~H. Hoang and G.-W. Kim, ``Imu augment tightly coupled lidar-visual-inertial odometry for agricultural environments,'' \emph{IEEE Robotics and Automation Letters}, vol.~9, \href{http://dx.doi.org/10.1109/LRA.2024.3440728}{DOI 10.1109/LRA.2024.3440728}, no.~10, pp. 8483--8490, 2024.

\bibitem{lee2024switch}
J.~Lee, ``Switch-slam: Switching-based lidar-inertial-visual slam for degenerate environments,'' \emph{IEEE Robotics and Automation Letters}, 2024.

\bibitem{10816047}
W.~Chen, ``P2d-do: Degeneracy optimization for lidar slam with point-to-distribution detection factors,'' \emph{IEEE Robotics and Automation Letters}, \href{http://dx.doi.org/10.1109/LRA.2024.3522839}{DOI 10.1109/LRA.2024.3522839}, pp. 1--8, 2024.

\bibitem{zhou2020lidar}
H.~Zhou, ``Lidar/uwb fusion based slam with anti-degeneration capability,'' \emph{IEEE Transactions on Vehicular Technology}, vol.~70, no.~1, pp. 820--830, 2020.

\bibitem{nubert2022learning}
J.~Nubert, ``Learning-based localizability estimation for robust lidar localization,'' in \emph{2022 IEEE/RSJ International Conference on Intelligent Robots and Systems (IROS)}, pp. 17--24.\hskip 1em plus 0.5em minus 0.4em\relax IEEE, 2022.

\bibitem{10994466}
Y.~Li and W.~Zhang, ``Anti-degeneracy scheme for lidar slam based on particle filter in geometry feature-less environments,'' \emph{IEEE Robotics and Automation Letters}, vol.~10, \href{http://dx.doi.org/10.1109/LRA.2025.3568569}{DOI 10.1109/LRA.2025.3568569}, no.~7, pp. 6784--6791, 2025.

\bibitem{tuna2023x}
T.~Tuna, ``X-icp: Localizability-aware lidar registration for robust localization in extreme environments,'' \emph{IEEE Transactions on Robotics}, 2023.

\bibitem{torroba2020pointnetkl}
I.~Torroba, ``Pointnetkl: Deep inference for gicp covariance estimation in bathymetric slam,'' \emph{IEEE Robotics and Automation Letters}, vol.~5, no.~3, pp. 4078--4085, 2020.

\bibitem{landry2019cello}
D.~Landry, ``Cello-3d: Estimating the covariance of icp in the real world,'' in \emph{2019 International Conference on Robotics and Automation (ICRA)}, pp. 8190--8196.\hskip 1em plus 0.5em minus 0.4em\relax IEEE, 2019.

\bibitem{botteghi2020reinforcement}
N.~Botteghi, ``Reinforcement learning helps slam: Learning to build maps,'' \emph{The International Archives of the Photogrammetry, Remote Sensing and Spatial Information Sciences}, vol.~43, pp. 329--335, 2020.

\bibitem{messikommer2024reinforcement}
N.~Messikommer, ``Reinforcement learning meets visual odometry,'' in \emph{European Conference on Computer Vision}, pp. 76--92.\hskip 1em plus 0.5em minus 0.4em\relax Springer, 2024.

\bibitem{lillicrap2019continuouscontroldeepreinforcement}
\BIBentryALTinterwordspacing
T.~P. Lillicrap, ``Continuous control with deep reinforcement learning,'' 2019. [Online]. Available: \url{https://arxiv.org/abs/1509.02971}
\BIBentrySTDinterwordspacing

\bibitem{christodoulou2019softactorcriticdiscreteaction}
\BIBentryALTinterwordspacing
P.~Christodoulou, ``Soft actor-critic for discrete action settings,'' 2019. [Online]. Available: \url{https://arxiv.org/abs/1910.07207}
\BIBentrySTDinterwordspacing

\bibitem{grupp2017evo}
M.~Grupp, ``evo: Python package for the evaluation of odometry and slam.'' \url{https://github.com/MichaelGrupp/evo}, 2017.

\end{thebibliography}

\end{document}